\documentclass[conference]{IEEEtran}
\IEEEoverridecommandlockouts
\usepackage{cite}
\usepackage{amsmath,amssymb,amsfonts}
\usepackage{graphicx}
\usepackage{textcomp}
\usepackage{xcolor}
\usepackage{booktabs}
\usepackage{url}
\makeatletter
\newcommand{\linebreakand}{%
  \end{@IEEEauthorhalign}
  \hfill\mbox{}\par
  \mbox{}\hfill\begin{@IEEEauthorhalign}
}
\makeatother

\begin{document}

\title{RGB-D SLAM on the Edge: Benchmarking and Profiling RTAB-Map and ORB-SLAM3 on the Raspberry Pi~5}

\author{\IEEEauthorblockN{Himanshu Patil}
\IEEEauthorblockA{\textit{Indian Institute of Technology Bombay}}
\and
\IEEEauthorblockN{Jaideep Joshi}
\IEEEauthorblockA{\textit{Indian Institute of Technology Bombay}}
\linebreakand
\IEEEauthorblockN{Rohit Saluja}
\IEEEauthorblockA{\textit{Indian Institute of Technology Mandi}}
\and
\IEEEauthorblockN{Ganesh Ramakrishnan}
\IEEEauthorblockA{\textit{Indian Institute of Technology Bombay}}
}

\maketitle

\begin{abstract}
Micro aerial vehicles that operate without GPS often run SLAM on ARM-based edge boards, yet there is little published data on how standard RGB-D SLAM systems behave on such hardware. We benchmark RTAB-Map and ORB-SLAM3 on the Raspberry Pi~5 using six TUM RGB-D sequences at three image resolutions, recording tracking coverage, trajectory error, time per frame, memory, CPU load and temperature for every run, and we profile both systems at the stage and function level. RTAB-Map was configured to process and keep every frame, so its back end runs on every frame rather than at its default rate of 1\,Hz. On the sequences that every configuration tracked, its front end takes 38--74\,ms per frame and its full pipeline 150--462\,ms, compared with 13--45\,ms and 26--62\,ms for ORB-SLAM3. RTAB-Map's main cost is optimizing a pose graph that keeps growing, because its working memory is unbounded by default; ORB-SLAM3's local bundle adjustment costs about as much as its tracking but runs in a parallel thread. At quarter resolution, both systems lose tracking on most sequences, and runs that lose tracking finish sooner and report lower error on the poses they keep, so time and accuracy must be read alongside coverage. Timing is repeatable to within 1--5\%, but trajectory error varies by up to 27\% between identical runs, so the accuracy differences between the two systems are inconclusive. No thermal throttling occurred.
\end{abstract}

\section{Introduction}
Simultaneous Localization and Mapping (SLAM) lets a robot build a map of an unknown environment while estimating its own pose within it. For micro-UAVs operating indoors without GPS, onboard SLAM is a necessity, and the computer that runs it is a primary design constraint: a tight payload budget for compute and sensing rules out desktop GPUs and high-power accelerators, leaving ARM-based edge boards as the practical choice.

RTAB-Map and ORB-SLAM3 are two widely used open-source RGB-D SLAM systems. ORB-SLAM3 \cite{Campos_2021} is among the most accurate open-source feature-based systems, with parallel tracking, local mapping and loop closing. RTAB-Map \cite{Labb__2018} offers configurable cost through its memory management and supports visual, RGB-D and inertial inputs through one interface, which makes it a common choice for embedded deployment. To our knowledge, the only peer-reviewed evaluation on the Raspberry Pi~5, whose Cortex-A76 cores are a large step over its predecessor, covers ORB-SLAM3 alone \cite{sud2026orbslam3pi5}; RTAB-Map has not been measured on the board, and the two systems have not been compared under one protocol.

This paper reports such measurements. It does not propose a new method; it provides:
\begin{itemize}
\item A benchmark of RTAB-Map \cite{Labb__2018} on the Raspberry Pi~5 at three image decimation levels over six TUM RGB-D sequences \cite{sturm12iros}, reporting tracking continuity, trajectory error, time per frame, memory, CPU load and temperature for every run, with recorded software provenance and repeatability measurements.
\item A stage-level and function-level profile of RTAB-Map on the Raspberry Pi~5 showing where per-frame time goes.
\item The same measurements and profiling for ORB-SLAM3 under the closest equivalent configuration, with a run-by-run comparison against RTAB-Map.
\end{itemize}

\section{Related Work}

\subsection{RGB-D SLAM Systems}
RGB-D SLAM has progressed from feature-based pipelines to scalable multi-map frameworks. Klein et al. \cite{4538852} established the template most modern systems follow: decoupling time-critical front-end pose estimation from heavier back-end refinement. RGB-D SLAM \cite{6594910} implemented this with g2o pose-graph optimization and an OctoMap volumetric map, but its cost grew with trajectory length as the map grew without bound. ORB-SLAM2 \cite{Mur_Artal_2017} extended the feature-based paradigm to RGB-D with robust relocalization and map reuse, and inherited the same unbounded map growth. Dense surfel-based approaches such as ElasticFusion \cite{doi:10.1177/0278364916669237} reconstruct surfaces in real time with deformation-graph loop closure, at memory and compute costs that make direct deployment on ARM boards impractical. Direct methods such as that of Kerl et al. \cite{6696650} skip feature extraction by minimizing photometric error, reducing front-end cost but adding sensitivity to illumination changes. RTAB-Map \cite{Labb__2018} introduced a memory management scheme that partitions the map into working, short-term and long-term memory so that loop closure detection can be held within a time budget regardless of map size, and supports visual, RGB-D and inertial inputs through one interface. ORB-SLAM3 \cite{Campos_2021} is among the most accurate open-source systems, with tight visual-inertial coupling and a multi-map atlas, but provides no mechanism for bounding its cost under resource constraints. RTAB-Map and ORB-SLAM3 are two widely used RGB-D systems, and represent two different designs: a graph that can retain every frame with memory management that can bound it, and a keyframe-based system with parallel tracking, local mapping and loop closing.

\subsection{Deployment of SLAM on Embedded Systems}
Deploying SLAM on embedded hardware is a balance between algorithmic cost and platform limits. Newcombe et al. \cite{inproceedings} showed dense real-time reconstruction on a desktop GPU, framing dense SLAM as a GPU-class problem. Delmerico et al. \cite{10.1109/ICRA.2018.8460664} evaluated visual-inertial odometry on embedded CPUs and showed that pipeline configuration, threading and hardware jointly determine real-time feasibility, which motivates configuration-level rather than system-level evaluation. Dedicated silicon such as the Navion ASIC \cite{suleiman2018navion2mwfullyintegrated} achieves extreme power efficiency at the cost of runtime configurability. Jetson modules moved the discussion toward GPU-assisted SLAM at UAV-relevant form factors \cite{GIUBILATO2019161, doi:10.2352/ISSN.2470-1173.2020.6.IRIACV-325}, yet configuration-level benchmarks of classical RGB-D SLAM on them remain sparse, and we are not aware of a peer-reviewed RTAB-Map characterization on the Orin Nano. Sud and Upadhyay \cite{sud2026orbslam3pi5} evaluated ORB-SLAM3 on the Raspberry Pi~5 across its sensor modes on TUM RGB-D, EuRoC and KITTI, reporting aggregate CPU, memory and execution time and finding no thermal throttling. They excluded RTAB-Map after informal tests indicated it could not run in real time on the board, and did not profile individual pipeline stages. Their RGB-D accuracy results differ from ours by two orders of magnitude while their resource measurements agree with ours; we discuss both in Section~\ref{sec:cmp}. Direct comparisons of the two systems' cost are few and depend on configuration. Labbé and Michaud \cite{Labb__2018} report, on a single desktop core, mean odometry times on TUM RGB-D of 70\,ms for RTAB-Map's F2M odometry against 54\,ms for ORB-SLAM2 used as RTAB-Map's front end, but 82\,ms against 175\,ms on KITTI. On a Jetson Xavier with stereo input, Abi Hana et al. \cite{abihana2026cuvslam} measure 60\,ms per frame for ORB-SLAM3 against 86\,ms for RTAB-Map's visual odometry, with RTAB-Map using less CPU (26\% against 36\%) and its SLAM back end updating at about 1\,Hz against 12.5\,Hz for ORB-SLAM3. RTAB-Map is thus typically lighter in CPU load because its back end runs at a low fixed rate, while its front end is not faster than ORB-SLAM's tracking. This paper adds the first RTAB-Map measurements on the Raspberry Pi~5, a comparison of the two systems on the same data, protocol and board, stage-level and function-level profiles of both, and an evaluation that reports tracking coverage for every run.

\subsection{Datasets}
The choice of dataset determines which aspects of SLAM performance are observable. The TUM RGB-D benchmark \cite{sturm12iros} provides synchronized RGB and depth sequences across several indoor scenes with motion-capture ground truth, and is the primary testbed for RGB-D SLAM. The EuRoC MAV dataset \cite{10.1177/0278364915620033} provides visual-inertial sequences recorded onboard a MAV, but has no depth channel. ICL-NUIM \cite{6907054} offers synthetic RGB-D sequences with perfect ground truth, which isolates algorithmic error from sensor noise but limits generalization to real sensors. We use TUM RGB-D: its sequences, recorded handheld and on a wheeled robot, cover a range of motion speeds, lighting and scene structure in indoor environments, and its wide use makes our results comparable with prior work.

\section{Platform and Software}
The Raspberry Pi~5 (Model B rev.~1.1) has a Broadcom BCM2712 SoC with four Cortex-A76 cores at 2.4\,GHz and 16\,GB of LPDDR4X memory, and was cooled by the official Active Cooler. It ran Ubuntu 24.04.4 LTS (kernel 6.8.0-1065-raspi) and RTAB-Map 0.23.13 from the ROS~2 Jazzy distribution (package build of 2 October 2026), linked against OpenCV 4.6.0, GTSAM 4.2 and g2o. All runs used RTAB-Map's standalone \texttt{rtabmap-rgbd\_dataset} tool, which runs odometry and SLAM in one process without ROS middleware. The CPU frequency governor was set to \texttt{performance}, and every core read 2400\,MHz before each run. Trajectory error was computed with \texttt{evo} 1.38.1 \cite{grupp2017evo}. ORB-SLAM3 was built from the official repository at commit \texttt{4452a3c} (version 1.0) with GCC 13.3, against the same OpenCV 4.6.0, Eigen 3.4.0, Boost 1.83 and Pangolin 0.9.5. It was built in \texttt{RelWithDebInfo} mode (\texttt{-O2}, with frame pointers for profiling), which omits the \texttt{-O3 -march=native} of its default release build, so its timings are a conservative estimate of what the board can deliver. Building with C++17 required replacing a \texttt{bool} increment in loop closing with the equivalent assignment; no other change affects its behaviour.

\section{Experimental Methodology}
\subsection{RTAB-Map}
RTAB-Map \cite{Labb__2018} is a graph-based SLAM system with appearance-based loop closure. Each processed frame becomes a node in a pose graph. Nodes first enter a short-term memory (STM) and then a working memory (WM) that is searched for loop closures and optimized. RTAB-Map can transfer nodes from WM to a long-term memory (LTM) to bound the size of WM, and with it the per-frame cost of loop closure detection and graph optimization. This transfer is triggered only when a time budget (\texttt{Rtabmap/TimeThr}) or a node budget (\texttt{Rtabmap/MemoryThr}) is exceeded. Both default to 0, meaning unlimited, and this study keeps the defaults, so WM is unbounded in every run; the logs confirm that no node was transferred to LTM in any run.

We run RTAB-Map in RGB-D mode with its frame-to-map (F2M) visual odometry, which detects up to 1000 GFTT corners per frame, describes them with ORB descriptors, matches them against a local feature map, estimates motion with PnP RANSAC (at least 20 inliers) and refines it with local bundle adjustment in g2o \cite{kummerle2011g2o}. If odometry fails on a frame, it is reset and the frame is not added to the map; a reset starts a new map session, which can split the trajectory. The SLAM back end detects loop closures with a bag-of-words model over up to 500 GFTT/ORB features per node, adds proximity links between nodes close in space, and optimizes the pose graph with GTSAM \cite{dellaert2012gtsam}, RTAB-Map's default optimizer.

\subsection{Configurations}
We evaluate three configurations that differ only in the decimation applied to the input images before odometry (\texttt{Odom/ImageDecimation}), listed in Table~\ref{tab:configs} with all shared parameters. A node is added for every frame (\texttt{RGBD/LinearUpdate} and \texttt{RGBD/AngularUpdate} set to 0) and every frame is processed (\texttt{Rtabmap/DetectionRate} set to 0), so cost is not reduced by skipping frames. RTAB-Map's defaults are lighter: its SLAM back end processes frames at 1\,Hz and adds a node only after 0.1\,m or 0.1\,rad of motion, while odometry still runs on every frame. These settings, inherited from our benchmark script, make RTAB-Map's back end process and keep every frame; they represent a per-frame worst case, not a tuned or default deployment, and the unlimited proximity depth and node-per-frame setting increase the size of the graph and the number of proximity links. The back-end costs reported here are therefore those of per-frame operation; the odometry costs apply to any deployment.

\begin{table}[t]
\centering
\caption{Evaluated RTAB-Map configurations and shared parameters. All 531 parameters not listed or overridden here ran at their 0.23.13 defaults; \texttt{Rtabmap/StatisticLogged} and \texttt{WorkingDirectory} were also set, for logging only.}
\label{tab:configs}
\footnotesize
\renewcommand{\arraystretch}{1.1}
\begin{tabular}{c c c}
\toprule
\textbf{ID} & \texttt{Odom/ImageDecimation} & \textbf{Odometry resolution} \\
\midrule
D1 & 1 & 640$\times$480 \\
D2 & 2 & 320$\times$240 \\
D4 & 4 & 160$\times$120 \\
\bottomrule
\end{tabular}\par
\vspace{2mm}
\scriptsize
\resizebox{\columnwidth}{!}{%
\begin{tabular}{l l l}
\toprule
\textbf{Shared parameter} & \textbf{Used} & \textbf{Default} \\
\midrule
\texttt{Rtabmap/DetectionRate} & 0 (every frame) & 1\,Hz \\
\texttt{RGBD/LinearUpdate} & 0 (node per frame) & 0.1\,m \\
\texttt{RGBD/AngularUpdate} & 0 (node per frame) & 0.1\,rad \\
\texttt{Mem/STMSize} & 30 & 10 \\
\texttt{Kp/MaxFeatures} & 500 & 500 \\
\texttt{OdomF2M/MaxSize} & 1000 & 2000 \\
\texttt{RGBD/ProximityBySpace} & true & true \\
\texttt{RGBD/ProximityMaxGraphDepth} & 0 (unlimited) & 50 \\
\texttt{Rtabmap/MaxRetrieved} & 2 & 2 \\
\texttt{Rtabmap/TimeThr}, \texttt{MemoryThr} & 0 (WM unbounded) & 0 \\
\texttt{Grid/CellSize} & 0.05\,m & 0.05\,m \\
\texttt{Grid/RangeMax} & 4.0\,m & 5.0\,m \\
\texttt{RGBD/Enabled} & true & true \\
\texttt{Optimizer/Strategy} & GTSAM & GTSAM \\
Depth scale (tool constant) & 5000 counts/m & --- \\
\midrule
\multicolumn{3}{l}{\emph{Defaults in effect (not overridden)}} \\
\texttt{Odom/Strategy} & F2M (0) & F2M \\
\texttt{Vis/FeatureType} & GFTT/ORB (8) & GFTT/ORB \\
\texttt{Vis/MaxFeatures} & 1000 & 1000 \\
\texttt{Vis/MinInliers} & 20 & 20 \\
\texttt{Vis/CorType} & feature matching & feature matching \\
\texttt{OdomF2M/BundleAdjustment} & g2o (1) & g2o \\
\texttt{Kp/DetectorStrategy} & GFTT/ORB (8) & GFTT/ORB \\
\texttt{Optimizer/Iterations} & 20 & 20 \\
\texttt{Mem/RehearsalSimilarity} & 0.6 & 0.6 \\
\texttt{RGBD/OptimizeMaxError} & 3.0 & 3.0 \\
Intrinsics, distortion & tool constants, none & --- \\
\bottomrule
\end{tabular}}
\end{table}

\subsection{ORB-SLAM3 Configuration}
ORB-SLAM3 \cite{Campos_2021} was run in RGB-D mode without IMU through its \texttt{rgbd\_tum} example, with loop closing and map merging enabled and its stock TUM settings (Table~\ref{tab:orbconfig}). To match the RTAB-Map runs as closely as possible, the example was modified to process every frame as fast as possible instead of pacing frames at the camera rate, to run without its viewer, and to decimate the RGB image (area interpolation) and depth image (nearest neighbour) by 1, 2 or 4 after loading them at full resolution, with the camera intrinsics scaled accordingly. RTAB-Map's dataset tool divides TUM depth by a fixed 5000 counts per metre for all cameras, so ORB-SLAM3 used the same divisor; its stock freiburg2 file uses 5208, with which ORB-SLAM3's error on fr2\_desk at D1 is 0.019\,m instead of 0.092\,m, with the same coverage and loop closures. A fixed divisor of 5000 may therefore inflate both systems' errors on fr2\_desk. Several differences remain: RTAB-Map decimates only the odometry input while ORB-SLAM3 decimates its whole pipeline, including place recognition; ORB-SLAM3 runs local mapping and loop closing in parallel threads, while RTAB-Map's dataset tool runs odometry and SLAM in sequence in one loop (in deployment they usually run concurrently, for example as separate ROS~2 nodes, so we report front-end and back-end times separately); RTAB-Map builds a local occupancy grid for every node, which ORB-SLAM3 does not; and, although both systems use ORB descriptors and a 1000-feature budget in the front end, ORB-SLAM3 detects FAST corners in an 8-level scale pyramid where RTAB-Map detects GFTT corners. Neither system corrects lens distortion on TUM: RTAB-Map's dataset tool builds a camera model with intrinsics only, and ORB-SLAM3 reads the distortion coefficients but applies them only in monocular mode. Both therefore treat the fr1 and fr2 images, which have non-negligible radial distortion, as ideal pinhole images, which affects absolute ATE but not the comparison between the two. ORB-SLAM3 also inserts keyframes selectively, whereas the RTAB-Map configuration adds a map node for every frame. Per-stage times come from ORB-SLAM3's built-in instrumentation (\texttt{REGISTER\_TIMES}); its local bundle adjustment series is logged with the keyframe-culling times by an upstream defect, so local BA is reported from the instrumentation's summary statistics. ORB-SLAM3's vocabulary load (7.8--8.1\,s) is excluded from its time per frame.

\begin{table}[t]
\centering
\caption{ORB-SLAM3 settings used (stock TUM files except where noted).}
\label{tab:orbconfig}
\scriptsize
\renewcommand{\arraystretch}{1.1}
\resizebox{\columnwidth}{!}{%
\begin{tabular}{l l}
\toprule
\textbf{Setting} & \textbf{Value} \\
\midrule
Sensor mode & RGB-D, no IMU (\texttt{rgbd\_tum}) \\
\texttt{ORBextractor.nFeatures} & 1000 \\
\texttt{ORBextractor.scaleFactor}, \texttt{nLevels} & 1.2, 8 \\
\texttt{ORBextractor.iniThFAST}, \texttt{minThFAST} & 20, 7 \\
\texttt{Stereo.ThDepth} & 40.0 (baselines) \\
\texttt{RGBD.DepthMapFactor} & 5000 (stock fr2: 5208) \\
\texttt{Stereo.b} (fr1 / fr2 / fr3) & 0.07732 / 0.0767 / 0.0747\,m \\
$f_x, f_y$ fr1 (D1) & 517.31, 516.47 \\
$f_x, f_y$ fr2 (D1) & 520.91, 521.01 \\
$f_x, f_y$ fr3 (D1) & 535.4, 539.2 \\
Intrinsics and image size at D2 / D4 & divided by 2 / 4 \\
Distortion coefficients & in file, not applied (RGB-D) \\
\texttt{Camera.fps}, \texttt{Camera.RGB} & 30 (unused), 1 \\
Loop closing, map merging & on \\
Viewer & off \\
Frame pacing & none (every frame, no sleep) \\
Vocabulary & \texttt{ORBvoc.txt} (stock) \\
\bottomrule
\end{tabular}}
\end{table}

\subsection{Dataset}
We use six TUM RGB-D sequences \cite{sturm12iros} spanning scene complexity, motion and duration (Table~\ref{tab:sequences}). RGB and depth images were paired with the TUM's associate tool using a maximum timestamp difference of 20\,ms. The fr1 sequences are short, fast-motion trajectories in textured desk environments. fr2\_desk is the longest at 2893 frames and exposes the growth of the map over extended operation. The fr3 sequences test degraded visual conditions: fr3\_nstx (nostructure\_notexture\_near\_withloop) has neither texture nor geometric structure and contains a loop, and fr3\_stx (structure\_notexture\_near) pairs strong geometric structure with little texture.

\begin{table}[t]
\centering
\caption{TUM RGB-D sequences and abbreviations used.}
\label{tab:sequences}
\footnotesize
\renewcommand{\arraystretch}{1.1}
\setlength{\tabcolsep}{3.5pt}
\resizebox{\columnwidth}{!}{%
\begin{tabular}{l l r r}
\toprule
\textbf{Abbrev.} & \textbf{Sequence} & \textbf{Dur. (s)} & \textbf{Frames} \\
\midrule
fr1\_xyz & freiburg1\_xyz & 30.9 & 792 \\
fr1\_desk & freiburg1\_desk & 23.4 & 573 \\
fr1\_room & freiburg1\_room & 48.9 & 1352 \\
fr2\_desk & freiburg2\_desk & 99.4 & 2893 \\
fr3\_nstx & freiburg3\_nostructure\_notexture\_near\_withloop & 37.7 & 1093 \\
fr3\_stx & freiburg3\_structure\_notexture\_near & 36.4 & 1054 \\
\bottomrule
\end{tabular}}
\end{table}

\subsection{Procedure and Metrics}
Each configuration of each system was run once per sequence (18 runs per system), one run at a time, with a 300\,s cooldown between runs. Two cases were repeated three times to measure run-to-run variation: D4 on fr2\_desk and D1 on fr1\_room. For each run we record:
\begin{itemize}
\item \emph{Tracking}: coverage, the fraction of frames with ground truth for which the run produced a pose (pose pairs), taken as the largest number of pose pairs any run achieved on the sequence. A run is classed as tracked if its coverage is at least 95\%. We also report RTAB-Map's odometry resets and ORB-SLAM3 runs that never initialized. Where ORB-SLAM3 opened more than one map, ATE and coverage are computed over the largest single-map segment, since the saved trajectory mixes coordinate frames.
\item \emph{Trajectory error}: absolute trajectory error (ATE) RMSE from \texttt{evo} against TUM ground truth after SE(3) alignment without scale correction, together with the number of estimated poses matched to ground truth (pose pairs).
\item \emph{Time per frame}: end-to-end wall-clock time of the run divided by its number of frames, including image loading. Per-stage times come from RTAB-Map's statistics logs.
\item \emph{Resources}: sampled once per second: system-wide CPU utilization, resident memory (RSS) of the RTAB-Map process, CPU temperature, CPU frequency, and the firmware throttling and under-voltage flags (\texttt{vcgencmd get\_throttled}).
\end{itemize}
The benchmark orchestration and the 1\,Hz logger ran alongside every run; they add a small constant load common to all runs.

\section{Results}
Table~\ref{tab:runs} lists every run. Because tracking failures change both time and error, we describe tracking first.

\begin{table*}[t]
\centering
\caption{All 18 RTAB-Map runs on the Raspberry Pi~5. Track: ok if coverage (pose pairs as a share of frames with ground truth) is at least 95\%. fr2\_desk ground truth covers 2174 of its 2893 frames. ``---'': no trajectory exported, or fewer than 50 poses.}
\label{tab:runs}
\footnotesize
\renewcommand{\arraystretch}{1.05}
\setlength{\tabcolsep}{5pt}
\begin{tabular}{l c r c r r r r r r}
\toprule
\textbf{Seq.} & \textbf{Cfg.} & \textbf{Resets} & \textbf{Track.} & \textbf{Pose pairs (cov.)} & \textbf{ATE (m)} & \textbf{ms/frame} & \textbf{RSS pk (MB)} & \textbf{CPU (\%)} & \textbf{T pk ($^\circ$C)} \\
\midrule
fr1\_xyz & D1 & 0 & ok & 790 (100\%) & 0.011 & 320.5 & 752 & 33.7 & 58.4 \\
 & D2 & 0 & ok & 790 (100\%) & 0.012 & 183.7 & 473 & 36.1 & 57.9 \\
 & D4 & 1 & ok & 789 (100\%) & 0.020 & 141.5 & 275 & 35.4 & 56.8 \\
\addlinespace[2pt]
fr1\_desk & D1 & 0 & ok & 572 (100\%) & 0.031 & 197.5 & 615 & 27.5 & 51.8 \\
 & D2 & 0 & ok & 572 (100\%) & 0.039 & 108.1 & 395 & 28.2 & 51.8 \\
 & D4 & 146 & lost & 426 (74\%) & 0.052 & 58.2 & 215 & 26.2 & 51.2 \\
\addlinespace[2pt]
fr1\_room & D1 & 0 & ok & 1352 (100\%) & 0.055 & 225.8 & 1243 & 29.0 & 53.5 \\
 & D2 & 0 & ok & 1352 (100\%) & 0.062 & 135.0 & 768 & 31.1 & 52.9 \\
 & D4 & 915 & lost & 0 (0\%) & --- & 45.0 & 214 & 27.4 & 51.8 \\
\addlinespace[2pt]
fr2\_desk & D1 & 0 & ok & 2174 (100\%) & 0.131 & 501.1 & 2295 & 38.2 & 56.2 \\
 & D2 & 0 & ok & 2174 (100\%) & 0.123 & 354.5 & 1615 & 38.7 & 56.2 \\
 & D4 & 0 & ok & 2174 (100\%) & 0.142 & 152.6 & 468 & 34.0 & 54.0 \\
\addlinespace[2pt]
fr3\_nstx & D1 & 749 & lost & 0 (0\%) & --- & 125.1 & 553 & 26.8 & 52.9 \\
 & D2 & 1 & lost & 0 (0\%) & --- & 176.3 & 1068 & 28.6 & 52.4 \\
 & D4 & 1068 & lost & 20 (2\%) & --- & 43.6 & 147 & 26.5 & 51.2 \\
\addlinespace[2pt]
fr3\_stx & D1 & 0 & ok & 1052 (100\%) & 0.068 & 257.5 & 1221 & 30.9 & 54.0 \\
 & D2 & 0 & ok & 1052 (100\%) & 0.036 & 168.7 & 803 & 29.9 & 52.9 \\
 & D4 & 911 & lost & 141 (13\%) & 0.031 & 49.2 & 181 & 26.3 & 52.9 \\
\bottomrule
\end{tabular}
\end{table*}

\subsection{Tracking Continuity}
By the coverage criterion, tracking held in 12 of the 18 runs. Quarter-resolution input (D4) lost tracking on four of the six sequences: fr1\_desk (146 resets, 74\% coverage), fr1\_room (915 resets, no trajectory), fr3\_nstx (1068 resets) and fr3\_stx (911 resets, 13\% coverage). It tracked fr1\_xyz, with one reset, and fr2\_desk, the slowest-moving sequence. No configuration tracked fr3\_nstx, the sequence with neither texture nor structure: D1 lost tracking (749 resets), and in D2 a single reset split the map so that no trajectory could be exported.

Tracking loss distorts the other metrics in a consistent direction. A run that loses tracking stops adding frames to the map, so it finishes sooner: D4 on fr3\_nstx took 43.6\,ms per frame against 176.3\,ms for D2, with its SLAM back end idle on almost every frame. Trajectory error is computed only over the poses that survive, which are the easy segments: D4 on fr3\_stx has the lowest ATE of the three configurations on that sequence (0.031\,m), scored on 141 of 1054 frames. Time and accuracy figures for SLAM on edge hardware therefore need to be reported together with tracking coverage; averaging them over runs that lost tracking would rank the failing configuration as both the fastest and the most accurate.

\subsection{Processing Time}
\begin{figure}[t]
\centering
\includegraphics[width=\columnwidth]{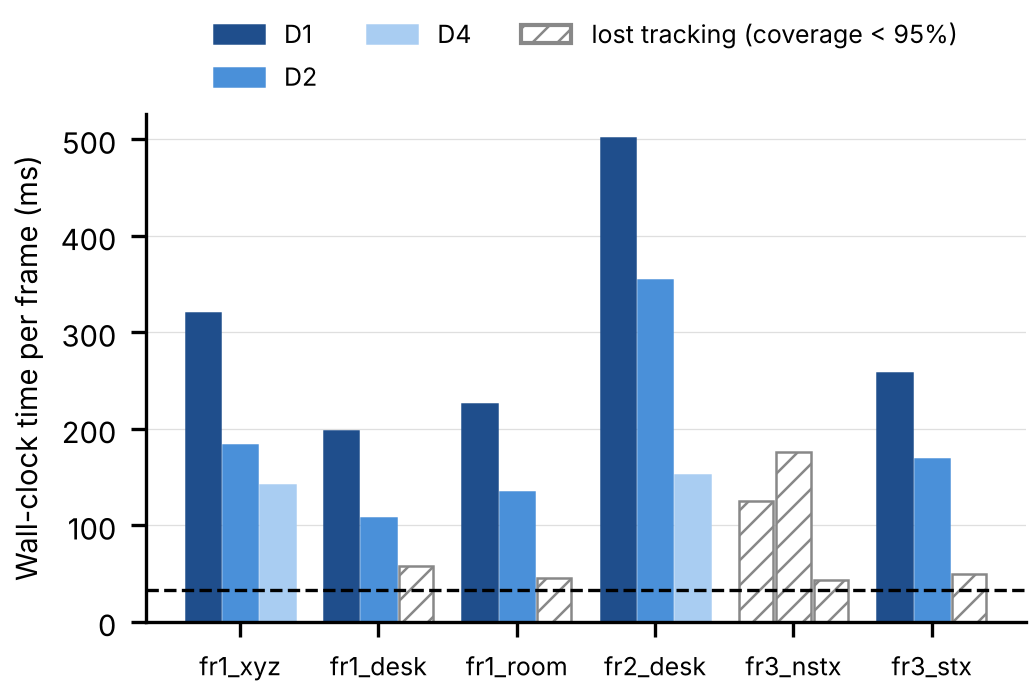}
\caption{Wall-clock time per frame for each run on the Raspberry Pi~5. Hatched bars mark runs that lost tracking; they finish sooner because frames without odometry are not processed by SLAM.}
\label{fig:seqtime}
\end{figure}

Fig.~\ref{fig:seqtime} shows time per frame for every run. On the two sequences that all three configurations tracked (fr1\_xyz and fr2\_desk), pooled time per frame (frame-weighted, with fr2\_desk contributing 79\% of the frames) is 462\,ms for D1, 318\,ms for D2 and 150\,ms for D4, a $3.1\times$ reduction from full to quarter resolution. Over all runs that tracked, D1 and D2 average 359 and 239\,ms per frame (five sequences each). Time per frame depends strongly on the sequence: with D1 it ranges from 198\,ms on fr1\_desk to 501\,ms on fr2\_desk, because cost grows with the number of nodes in the map (Section~\ref{sec:profiling}). No configuration approaches the 33.3\,ms frame interval of the 30\,fps sequences; the fastest run that tracked, D2 on fr1\_desk, took 108\,ms per frame. The repeated runs reproduce time per frame closely: 160.7\,$\pm$\,7.1\,ms for D4 on fr2\_desk and 218.8\,$\pm$\,6.2\,ms for D1 on fr1\_room (mean\,$\pm$\,standard deviation over three runs, a coefficient of variation of 4.4\% and 2.8\%).

\subsection{Trajectory Accuracy}
Among runs with full tracking, ATE RMSE ranges from 0.011\,m on fr1\_xyz to 0.142\,m on fr2\_desk (Table~\ref{tab:runs}). The sequence dominates: on fr1\_xyz all three configurations stay within 0.020\,m, while on fr2\_desk all three lie between 0.123 and 0.142\,m. On the fr1 sequences error rises from D1 to D2 (and to D4 on fr1\_xyz), whereas on fr2\_desk and fr3\_stx D2 is lower than D1; differences of this size are of the same order as the run-to-run variation. Three identical runs of D1 on fr1\_room gave 0.050, 0.055 and 0.081\,m (coefficient of variation 27\%), and three of D4 on fr2\_desk gave 0.123, 0.126 and 0.142\,m (8\%). RANSAC sampling and multi-threaded optimization make RTAB-Map non-deterministic. With a coefficient of variation of up to 27\% per run, the difference between two single runs has a spread of about 38\%, so single-run differences in ATE of less than about 40\% should not be read as real.

\subsection{Resource Use and Thermal Behavior}
RTAB-Map keeps little more than one core busy: mean system-wide CPU utilization was 26--39\% across runs on the four-core board. The resident memory of the RTAB-Map process grows with the map and falls with decimation; on fr2\_desk it peaked at 2295, 1615 and 468\,MB for D1, D2 and D4. Peak CPU temperature stayed between 51.2 and 58.4\,$^\circ$C, the CPU frequency remained at 2400\,MHz in every sample of every run, and no thermal throttling occurred. The firmware flagged supply under-voltage in 3 of 1144 samples during D1 on fr2\_desk, without any drop in frequency; this indicates a marginal power supply rather than a thermal limit.

\section{Computational Profiling}\label{sec:profiling}
This section examines where the time goes in each system for D4 on fr2\_desk, the longest sequence that every RTAB-Map configuration tracked.

\subsection{RTAB-Map}
For RTAB-Map, wall-clock time per stage comes from RTAB-Map's per-iteration statistics logs of the benchmark run. Cycle shares per function come from a separate run under Linux \texttt{perf}, sampling with frame-pointer call graphs and RTAB-Map debug symbols; they attribute cost within the pipeline and are not used as wall-clock data.

\begin{table}[t]
\centering
\caption{Mean per-frame computation time by component, D4 on fr2\_desk (2893 frames). Share is of odometry plus SLAM (135.6\,ms); the remaining 17.0\,ms of the 152.6\,ms wall-clock time per frame is spent outside both, mostly loading and decoding images.}
\label{tab:profile}
\footnotesize
\renewcommand{\arraystretch}{1.1}
\setlength{\tabcolsep}{2.2pt}
\begin{tabular}{l r r r r r}
\toprule
\textbf{Component} & \textbf{Mean} & \textbf{p50} & \textbf{p95} & \textbf{Max} & \textbf{Share} \\
\midrule
Visual odometry (front end) & 44.9 & 48.0 & 60.0 & 68.0 & 33.1\% \\
\midrule
Graph optimization$^{a}$ & 38.6 & & & & 28.4\% \\
Other (untimed)$^{b}$ & 38.4 & & & & 28.3\% \\
Memory update & 7.8 & 8.1 & 9.4 & 12.2 & 5.8\% \\
Likelihood, retrieval, stats & 5.2 & 5.9 & 9.7 & 15.8 & 3.9\% \\
Bayes filter posterior & 0.7 & 0.7 & 1.2 & 1.8 & 0.5\% \\
SLAM back end subtotal & 90.7 & 60.0 & 268.0 & 520.0 & 66.9\% \\
\midrule
Total (odometry + SLAM) & 135.6 & & & & 100\% \\
\bottomrule
\end{tabular}

\vspace{1mm}
\raggedright\scriptsize $^{a}$RTAB-Map's logs give only its mean. Graph optimization and the in-timer part of other combined: p50 30.4, p95 227.0, max 478.8\,ms.\\
$^{b}$27.7\,ms inside RTAB-Map's per-iteration timer not assigned to any stage, plus 10.7\,ms of the SLAM step outside that timer.
\end{table}

\begin{figure}[t]
\centering
\includegraphics[width=\columnwidth]{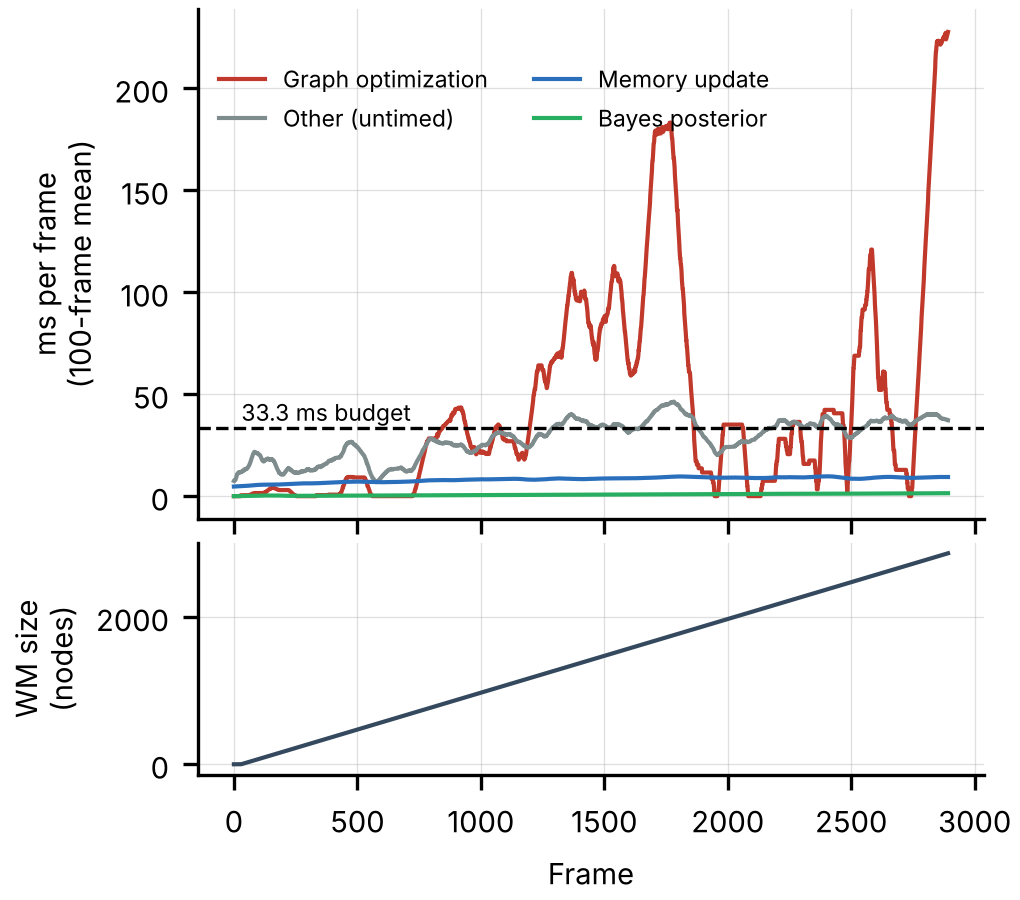}
\caption{Per-stage SLAM time over the run (100-frame moving mean) and working-memory size for D4 on fr2\_desk. Shown from a run with identical parameters under the \texttt{ondemand} governor, whose mean SLAM time is 9\% higher than that of the benchmark run in Table~\ref{tab:profile}. The dashed line is the 33.3\,ms frame interval.}
\label{fig:profile}
\end{figure}

\textbf{Working memory is never bounded.} No node was transferred to long-term memory, and working memory grew by one node per frame to 2863 nodes (Fig.~\ref{fig:profile}, bottom). Stages that depend on map size grow from the first third of the sequence to the second and stay high: mean SLAM time per frame, as measured by RTAB-Map's internal timer (which excludes the 10.7\,ms of the SLAM step outside it, Table~\ref{tab:profile}), is 34.9, 114.3 and 90.8\,ms in the three thirds. The Bayes filter posterior tracks working-memory size almost exactly (Spearman correlation 0.99), but at this map size stays below 2\,ms.

\textbf{Graph optimization is the largest cost.} Global pose-graph optimization averages 38.6\,ms per frame, 28.4\% of all computation and 42.5\% of SLAM time, rising from 9.1\,ms in the first third to 65.8\,ms in the second and 40.8\,ms in the last. It is near zero on most frames and runs in bursts when proximity or loop-closure links are added to the graph, which the unlimited proximity depth and node-per-frame settings make frequent (Fig.~\ref{fig:profile}): graph optimization and the untimed work inside RTAB-Map's timer combined have a median of 30\,ms but a 95th percentile of 227\,ms and a maximum of 479\,ms. \texttt{perf} attributes 19.7\% of all cycles to \texttt{Rtabmap::optimizeCurrentMap}, most of it inside GTSAM, with Cholesky elimination (10.7\%) and Gauss-Newton iteration (10.3\%) the largest parts. RTAB-Map's timers do not attribute the remaining other work; computing proximity-link transforms (\texttt{Memory::computeTransform}, 8.2\% of cycles) and generating each node's local occupancy grid are likely contributors, which we have not separated. In total, 71.3\% of frames exceed the 33.3\,ms interval in the SLAM back end alone.

\textbf{Odometry.} Visual odometry, RTAB-Map's front end and the source of the per-frame pose estimate, takes 44.9\,ms per frame on average and at most 68\,ms in this run, so at quarter resolution it alone exceeds the frame interval. Its cost does not shrink in proportion to resolution: on fr2\_desk mean odometry time is 75.5, 38.2 and 44.9\,ms for D1, D2 and D4, against SLAM back-end means of 395.1, 298.6 and 90.7\,ms, so the back end dominates at every resolution and odometry accounts for 16\%, 11\% and 33\% of their sum. \texttt{perf} attributes 24.1\% of cycles to \texttt{Odometry::process}, 23.3\% of them within F2M motion estimation. The registration, PnP RANSAC and bundle-adjustment routines it uses are also called by the SLAM back end to compute proximity and loop-closure links (\texttt{Memory::computeTransform}, 8.2\%), so their inclusive shares cannot be assigned to odometry alone, and RTAB-Map's statistics logs do not time the odometry stages separately. A per-stage breakdown of odometry is therefore not part of this profile.

\textbf{Runtime overhead.} Memory allocation appears in 13.2\% of sampled call stacks (\texttt{malloc}), and Intel TBB, used by GTSAM for parallel elimination, in 25.3\%. PNG decoding of the dataset images costs about 5\% of cycles, a cost a live camera would not incur. The SLAM back end accounts for 46.0\% of cycles against 24.1\% for odometry, consistent with the stage times.

\textbf{Summary.} In this configuration the dominant SLAM cost scales with the size of working memory, which RTAB-Map bounds only when \texttt{Rtabmap/MemoryThr} or \texttt{Rtabmap/TimeThr} is set; both were left at their defaults (unbounded) throughout this study. The SLAM back end costs more than odometry at every resolution.

\subsection{ORB-SLAM3}
For ORB-SLAM3, stage times come from its built-in instrumentation in the benchmark run, and cycle shares from a separate \texttt{perf} run with frame-pointer call graphs, split by the thread each sample belongs to. The profiled run used ORB-SLAM3's stock freiburg2 depth factor; its time per frame (26.7\,ms) is within 2\% of the run with RTAB-Map's depth factor (26.2\,ms). Unlike RTAB-Map, ORB-SLAM3 runs tracking, local mapping and loop closing in three concurrent threads, so stage times overlap in wall-clock time; Table~\ref{tab:orbprofile} therefore gives shares of summed thread time (79.6\,s over the run).

\begin{table}[t]
\centering
\caption{ORB-SLAM3 computation time by component (ms), D4 on fr2\_desk. Tracking rows are per frame (2893 frames); local-mapping rows are per local-mapping iteration (412). Share is of summed busy time of all three threads.}
\label{tab:orbprofile}
\footnotesize
\renewcommand{\arraystretch}{1.1}
\setlength{\tabcolsep}{1.9pt}
\begin{tabular}{l r r r r r}
\toprule
\textbf{Component} & \textbf{Mean} & \textbf{p50} & \textbf{p95} & \textbf{Max} & \textbf{Share} \\
\midrule
\multicolumn{6}{l}{\emph{Tracking thread (front end), per frame}} \\
ORB extraction & 3.2 & 3.3 & 4.3 & 6.2 & 11.4\% \\
Pose prediction & 3.0 & 2.8 & 4.9 & 7.4 & 10.8\% \\
Local-map tracking & 6.4 & 5.7 & 12.1 & 16.3 & 23.0\% \\
Keyframe decision & 0.1 & 0.1 & 0.3 & 0.6 & 0.3\% \\
Other (untimed) & 1.1 & & & & 4.2\% \\
Tracking subtotal & 13.7 & 13.4 & 22.0 & 27.4 & 49.8\% \\
\midrule
\multicolumn{6}{l}{\emph{Local-mapping thread, per iteration}} \\
Keyframe insertion & 4.6 & 4.3 & 7.6 & 10.9 & 2.4\% \\
Map-point creation & 12.5 & 11.9 & 24.4 & 33.2 & 6.5\% \\
Map-point culling & 0.1 & 0.1 & 0.2 & 0.3 & 0.0\% \\
Local BA and keyframe culling$^{a}$ & 77.7 & 65.2 & 212.9 & 275.9 & 40.2\% \\
Local-mapping subtotal & 94.9 & 80.9 & 244.9 & 310.2 & 49.1\% \\
\midrule
\multicolumn{6}{l}{\emph{Loop-closing thread}} \\
Place recognition$^{b}$ & 2.2 & & & & 1.1\% \\
\bottomrule
\end{tabular}

\vspace{1mm}
\raggedright\scriptsize $^{a}$Local BA alone: 84.4\,$\pm$\,61.8\,ms per execution over 348 executions (107 aborted); keyframe culling 7.6\,ms. ORB-SLAM3's instrumentation logs only the mean of each separately.\\
$^{b}$Per keyframe, mean only; no loop was closed in this run.

\end{table}

\textbf{Tracking and local mapping cost the same.} The tracking thread was busy for 39.7\,s and the local-mapping thread for 39.1\,s of the 77.3\,s of processing after the 7.9\,s vocabulary load (51.3\% and 50.6\%), and loop closing for about 0.9\,s, an estimate from its mean cost per keyframe. Local mapping is therefore half of ORB-SLAM3's computation but does not appear in its per-frame time, because it runs on another core. Within tracking, local-map tracking (6.4\,ms per frame) costs about as much as ORB extraction and pose prediction together. Within local mapping, local bundle adjustment dominates: it averages 84.4\,ms per execution, and with keyframe culling accounts for 82\% of local-mapping time. Local bundle adjustment optimizes the new keyframe, its covisible keyframes and the points they observe; ORB-SLAM3 runs a global bundle adjustment over the whole map only after a loop closure or map merge, and none occurred in this run.

\begin{table}[t]
\centering
\caption{ORB-SLAM3 share of CPU cycles by thread and function (\texttt{perf}), D4 on fr2\_desk.}
\label{tab:orbperf}
\footnotesize
\renewcommand{\arraystretch}{1.1}
\begin{tabular}{l r}
\toprule
\textbf{Thread / function} & \textbf{Cycles} \\
\midrule
Local mapping & 32.0\% \\
\quad local bundle adjustment & 23.5\% \\
\quad neighbour search and fusion & 3.9\% \\
\quad other & 4.6\% \\
Tracking & 30.9\% \\
\quad descriptor matching & 11.6\% \\
\quad ORB extraction & 6.9\% \\
\quad pose optimization & 5.7\% \\
\quad other & 6.7\% \\
Loop closing & 0.8\% \\
Main thread outside tracking & 36.4\% \\
\quad PNG decoding & 25.3\% \\
\quad vocabulary loading & 5.7\% \\
\quad other & 5.5\% \\
\bottomrule
\end{tabular}
\end{table}

\textbf{Cycle shares agree.} \texttt{perf} gives the same split between the two main threads (Table~\ref{tab:orbperf}): 32.0\% of cycles in local mapping and 30.9\% in tracking. Local bundle adjustment is the largest single function at 23.5\% of all cycles, 69\% of it inside g2o's sparse Levenberg-Marquardt solver. In tracking, descriptor matching against the local map takes more cycles (11.6\%) than ORB extraction (6.9\%) or pose optimization (5.7\%), so the front end is bound by feature association rather than detection. A quarter of all cycles go to decoding the dataset PNG images in the main thread, which accounts for most of the 13.0\,ms per frame between ORB-SLAM3's tracking time and its time per frame; a live camera would deliver decoded frames.

\subsection{Side by Side}
\begin{figure}[t]
\centering
\includegraphics[width=\columnwidth]{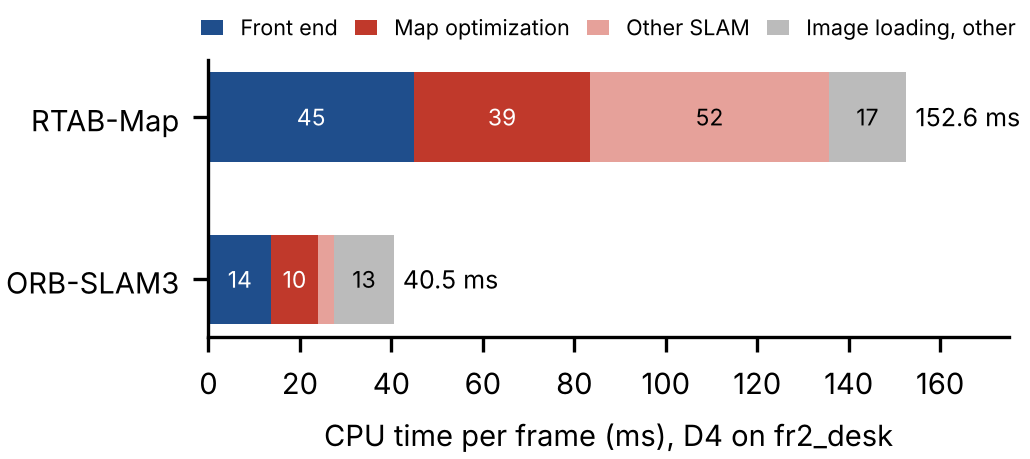}
\caption{CPU time per frame by component, D4 on fr2\_desk. ORB-SLAM3 runs local mapping and loop closing (included in map optimization and other SLAM) concurrently with tracking, so its time per frame is 26.7\,ms. RTAB-Map's components run in sequence in the dataset tool.}
\label{fig:breakdown}
\end{figure}

Fig.~\ref{fig:breakdown} sets the two breakdowns next to each other as CPU time per frame. RTAB-Map spends 153\,ms per frame, all of it in sequence in the dataset tool; ORB-SLAM3 spends 40.5\,ms of CPU time per frame, of which 26.7\,ms is on the critical path because local mapping runs concurrently. The front end differs by $3.3\times$ (44.9 against 13.7\,ms) and map optimization by $3.8\times$ (38.6\,ms against 10.2\,ms per frame). The two optimize different problems: RTAB-Map optimizes a pose graph over all nodes, poses only, and is expensive when proximity or loop-closure links are added, whereas ORB-SLAM3 refines a local window of keyframes and their points and reserves global optimization for loop events. The largest gap is in the rest of the SLAM back end (52.1 against 3.6\,ms), most of which RTAB-Map's stage timers do not attribute; it grows with the number of map nodes, and in this configuration RTAB-Map adds 2863 nodes where ORB-SLAM3 keeps about 215 keyframes. Image loading costs 13--17\,ms per frame in both.

\section{Comparison with ORB-SLAM3}\label{sec:cmp}
Table~\ref{tab:cmp} compares the two systems run by run.

\begin{table*}[t]
\centering
\caption{RTAB-Map / ORB-SLAM3, run by run, on the Raspberry Pi~5. Track: ok if coverage is at least 95\%, lost, or no init (ORB-SLAM3 never initialized). Front end: RTAB-Map odometry mean / ORB-SLAM3 tracking (\texttt{TrackRGBD}) mean. ``---'': no trajectory, or fewer than 50 poses.}
\label{tab:cmp}
\scriptsize
\renewcommand{\arraystretch}{1.05}
\setlength{\tabcolsep}{5pt}
\begin{tabular}{l c c c c c c c}
\toprule
\textbf{Seq.} & \textbf{Cfg.} & \textbf{Track} & \textbf{Pose pairs (cov.)} & \textbf{ATE (m)} & \textbf{ms/frame} & \textbf{Front end (ms)} & \textbf{RSS peak (MB)} \\
\midrule
fr1\_xyz & D1 & ok / ok & 790 / 790 & 0.011 / 0.010 & 321 / 64 & 70 / 44 & 752 / 567 \\
 & D2 & ok / ok & 790 / 790 & 0.012 / 0.011 & 184 / 39 & 37 / 27 & 473 / 543 \\
 & D4 & ok / ok & 789 / 790 & 0.020 / 0.022 & 141 / 26 & 32 / 13 & 275 / 521 \\
\addlinespace[2pt]
fr1\_desk & D1 & ok / ok & 572 / 571 & 0.031 / 0.024 & 197 / 63 & 63 / 43 & 615 / 598 \\
 & D2 & ok / ok & 572 / 572 & 0.039 / 0.023 & 108 / 42 & 31 / 29 & 395 / 555 \\
 & D4 & lost / lost & 426 / 532 & 0.052 / 0.129 & 58 / 29 & 26 / 16 & 215 / 535 \\
\addlinespace[2pt]
fr1\_room & D1 & ok / ok & 1352 / 1352 & 0.055 / 0.066 & 226 / 61 & 65 / 41 & 1243 / 790 \\
 & D2 & ok / ok & 1352 / 1352 & 0.062 / 0.091 & 135 / 35 & 31 / 23 & 768 / 708 \\
 & D4 & lost / lost & 0 / 1242 & --- / 1.051 & 45 / 23 & 24 / 10 & 214 / 567 \\
\addlinespace[2pt]
fr2\_desk & D1 & ok / ok & 2174 / 2174 & 0.131 / 0.092 & 501 / 61 & 75 / 45 & 2295 / 755 \\
 & D2 & ok / ok & 2174 / 2174 & 0.123 / 0.110 & 355 / 43 & 38 / 30 & 1615 / 751 \\
 & D4 & ok / lost & 2174 / 1646 & 0.142 / 0.281$^{\dagger}$ & 153 / 26 & 45 / 13 & 468 / 611 \\
\addlinespace[2pt]
fr3\_nstx & D1 & lost / lost & 0 / 0 & --- / --- & 125 / 35 & 69 / 16 & 553 / 537 \\
 & D2 & lost / no init & 0 / 0 & --- / --- & 176 / 14 & 46 / 3 & 1068 / 491 \\
 & D4 & lost / no init & 20 / 0 & --- / --- & 44 / 13 & 30 / 1 & 147 / 490 \\
\addlinespace[2pt]
fr3\_stx & D1 & ok / lost & 1052 / 534 & 0.068 / 0.017 & 257 / 35 & 64 / 17 & 1221 / 569 \\
 & D2 & ok / no init & 1052 / 0 & 0.036 / --- & 169 / 14 & 46 / 3 & 803 / 491 \\
 & D4 & lost / no init & 141 / 0 & 0.031 / --- & 49 / 13 & 32 / 1 & 181 / 490 \\
\bottomrule
\end{tabular}

\vspace{1mm}
\raggedright\scriptsize $^{\dagger}$ORB-SLAM3 opened a second map; ATE over the largest single-map segment.
\end{table*}

\textbf{Tracking.} By the coverage criterion, ORB-SLAM3 tracked 9 of 18 runs against 12 for RTAB-Map, and the two fail in different places. ORB-SLAM3 never initialized on either textureless fr3 sequence at half or quarter resolution, where RTAB-Map tracked fr3\_stx at every resolution except D4, and at D1 it initialized late on fr3\_stx, covering 51\% of the frames. At D4, ORB-SLAM3 lost tracking on fr1\_desk, fr1\_room and fr2\_desk but kept a trajectory on all three (93\%, 92\% and 76\% coverage, opening a second map on fr2\_desk), whereas RTAB-Map lost fr1\_desk and fr1\_room, exporting no trajectory at all on fr1\_room, and tracked fr2\_desk.

\begin{figure}[t]
\centering
\includegraphics[width=\columnwidth]{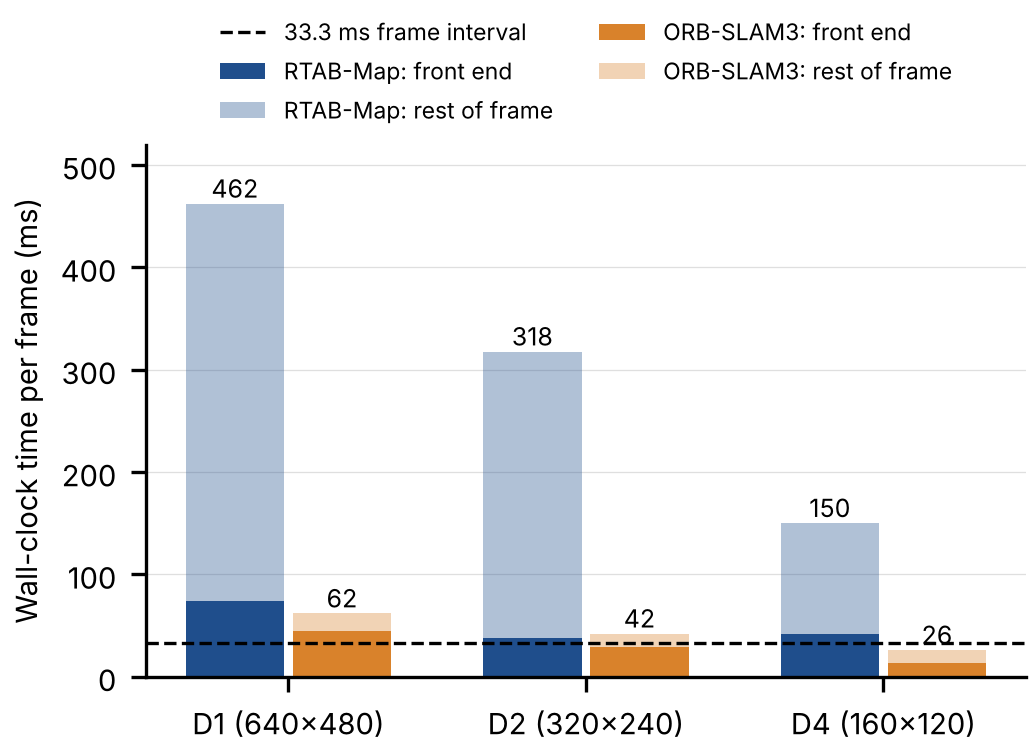}
\caption{Wall-clock time per frame pooled over fr1\_xyz and fr2\_desk, split into the front end (RTAB-Map odometry, ORB-SLAM3 tracking) and the rest of the frame. ORB-SLAM3's local mapping and loop closing run in parallel threads and are not part of its per-frame wall-clock time.}
\label{fig:cmp}
\end{figure}

\textbf{Front end and time per frame.} The front end delivers the per-frame pose a controller uses, and runs on every frame in any RTAB-Map deployment. Pooled over fr1\_xyz and fr2\_desk, ORB-SLAM3's tracking takes 44.7, 29.4 and 13.1\,ms per frame for D1, D2 and D4 against 74.3, 37.8 and 42.1\,ms for RTAB-Map's odometry, $1.7\times$, $1.3\times$ and $3.2\times$ less. End to end, ORB-SLAM3 takes 61.9, 42.2 and 26.1\,ms per frame against 462.3, 317.8 and 150.2\,ms for RTAB-Map in the dataset tool (Fig.~\ref{fig:cmp}). These pooled figures are frame-weighted and dominated by fr2\_desk; per sequence the ratio is $4.7$--$5.5\times$ on fr1\_xyz and $5.8$--$8.3\times$ on fr2\_desk. Because the dataset tool runs RTAB-Map's odometry and SLAM in sequence, this overstates the gap for a deployment in which they run concurrently; there, throughput is bounded by the slower stage, RTAB-Map's per-frame SLAM back end, at 356.9, 261.9 and 90.9\,ms, still $5.8\times$, $6.2\times$ and $3.5\times$ ORB-SLAM3's time per frame. At RTAB-Map's default back-end rate of 1\,Hz, which we did not measure, the bound would be its odometry. ORB-SLAM3 at D4 is the only configuration of either system within the 33.3\,ms frame interval on these sequences, including image loading, but on fr2\_desk it lost tracking and its largest map covers 76\% of the frames; on fr1\_xyz, which it tracked fully, it took 25.7\,ms. Further factors limit how far these differences reflect the algorithms rather than the configuration: RTAB-Map adds a node for every frame (2863 on fr2\_desk) where ORB-SLAM3 keeps about 215 keyframes; ORB-SLAM3 uses more of the board, with mean CPU utilization of 33--42\% against 26--39\%; ORB-SLAM3 was built at \texttt{-O2} without \texttt{-march=native}, and we did not establish the optimization level of the RTAB-Map binary package; and the firmware flagged supply under-voltage on 8--21\% of samples in every ORB-SLAM3 D1 run, with no change in reported CPU frequency, while a control rerun of the most affected run on the same supply was 4.7\% faster and showed no flags, so the D1 times carry an uncertainty of a few percent.

\textbf{Where the time goes.} Section~\ref{sec:profiling}-B and Fig.~\ref{fig:breakdown} break ORB-SLAM3's cost down by stage and thread. No loop was closed at D4 on fr2\_desk, against one at D1 and D2.

\textbf{Accuracy.} Among the nine runs that both systems tracked, ORB-SLAM3 has the lower ATE on six (fr1\_xyz D1 and D2, fr1\_desk D1 and D2, fr2\_desk D1 and D2) and RTAB-Map on three (fr1\_xyz D4, fr1\_room D1 and D2). The higher error exceeds the lower by more than 40\%, about the spread between two single runs, only on fr1\_desk D2 (73\%), fr1\_room D2 (47\%) and fr2\_desk D1 (43\%), so with one run per configuration the accuracy comparison is inconclusive. On fr3\_stx at D1, ORB-SLAM3's trajectory covers 51\% of the frames against 100\% for RTAB-Map, so the two errors are not comparable. ORB-SLAM3's accuracy falls sharply at D4 (0.281\,m over the largest map on fr2\_desk against 0.142\,m for RTAB-Map, and 1.05\,m on fr1\_room). Two properties of the configuration may contribute. Decimation reaches ORB-SLAM3's whole pipeline, and no loop was closed at D4 on fr2\_desk against one at D1 and D2. And with the stock 8-level pyramid at scale factor 1.2, the coarsest levels at D4 are about $45\times34$ pixels, against a border of about 19 pixels that ORB extraction leaves on each side, so they contribute few features. We kept the stock pyramid, since scaling it with resolution would be tuning.

\textbf{Relation to prior Raspberry Pi~5 measurements.} Sud and Upadhyay \cite{sud2026orbslam3pi5} report ORB-SLAM3 RGB-D results on TUM that differ substantially from ours: they report ATE of 2.0--3.0\,m with failure rates of 0.95--0.99 and 2--6 frames per second on fr1\_xyz, fr1\_desk, fr1\_room and fr2\_desk, whereas we measure 0.010--0.092\,m at full resolution with nearly every frame tracked. Our values are of the same order as those reported for ORB-SLAM2 on desktop hardware \cite{Mur_Artal_2017}, which suggests the difference lies in evaluation rather than in the platform. Their failure rates and frame rates are close to what scoring only keyframes against TUM's 100\,Hz motion-capture ground truth would give; with our keyframe counts this yields failure rates of 0.98, 0.97 and 0.95 on fr1\_xyz, fr1\_desk and fr1\_room against their 0.98, 0.96 and 0.95. Their resource measurements, by contrast, agree with ours: on the four fr1 and fr2 sequences they report 1.7--1.9 busy cores and 0.52--0.73\,GiB peak memory for RGB-D mode, against 1.8--1.9 cores (system-wide utilization times four) and 567--790\,MB in our full-resolution runs, while our execution times are 22--32\% longer, consistent with our build omitting release optimizations.

\textbf{Memory and repeatability.} ORB-SLAM3's process memory peaked at 490--790\,MB. Its floor of about 490\,MB is set by the vocabulary it loads at startup, but it grows far less with the map than RTAB-Map's: on fr2\_desk at D1 it peaked at 755\,MB against 2295\,MB. It is therefore lower than RTAB-Map at D1 on every sequence and higher at D4 on every sequence. Peak temperature was at most 59.0\,$^\circ$C. Three identical ORB-SLAM3 runs reproduced time per frame within 0.9\% (D4, fr2\_desk) and 5.2\% (D1, fr1\_room) and ATE within 1.7\% (D1, fr1\_room), but ATE varied by 14\% on D4 fr2\_desk, where one of the three runs lost tracking and opened a second map.

\section{Limitations and Future Work}
Each configuration was run once per sequence, apart from the two repeated cases, and those show that ATE varies by up to 27\% between identical runs; differences in ATE between configurations and systems are therefore indicative only, and repeated runs with medians would be needed to establish them. The RTAB-Map configuration processes and keeps every frame and the dataset tool runs its stages in sequence, so its back-end cost and end-to-end time per frame are those of a per-frame worst case; a default configuration with concurrent odometry and SLAM was not measured. ORB-SLAM3 was run with its stock pyramid at every decimation level. Both systems divided freiburg2 depth by 5000. Neither system corrected lens distortion, so absolute ATE values on fr1 and fr2 are not directly comparable with published results obtained with distortion correction. The ORB-SLAM3 comparison uses the closest available equivalent configuration, but the two systems differ in how decimation is applied, in threading and in keyframe selection, and ORB-SLAM3 was built without its release optimizations; its time per frame does not include the work of its parallel mapping threads. Measurements cover one RTAB-Map version and its default, unbounded working memory; measuring the same configurations with \texttt{Rtabmap/MemoryThr} or \texttt{Rtabmap/TimeThr} set would characterize the bounded mode, and repeating the methodology on the NVIDIA Jetson Orin Nano, including GPU-accelerated SLAM pipelines, would show how these results carry over to GPU-equipped edge boards.

\section{Conclusion}
We measured RTAB-Map~0.23.13 and ORB-SLAM3~1.0 on the Raspberry Pi~5 at three decimation levels over six TUM RGB-D sequences. On the sequences that all configurations tracked, RTAB-Map's odometry takes 38--74\,ms per frame and ORB-SLAM3's tracking 13--45\,ms. Configured to process and keep every frame, RTAB-Map takes 150--462\,ms per frame end to end in its dataset tool, against 26--62\,ms for ORB-SLAM3, whose local mapping, dominated by local bundle adjustment, costs as much as its tracking but runs in a parallel thread. With its default memory thresholds, RTAB-Map's working memory grows without bound and pose-graph optimization over the growing graph is its largest back-end cost. Quarter resolution breaks tracking for both systems on most sequences, the two fail on different sequences, and runs that lose tracking appear both faster and more accurate, so these metrics must be read with tracking coverage. Timing is reproducible to within 1--5\%, trajectory error only to within 27\%, so single-run accuracy comparisons are inconclusive. No thermal throttling occurred.

\bibliographystyle{IEEEtran}
\bibliography{references}
\end{document}